\documentclass[10pt,journal,compsoc]{IEEEtran}

\ifCLASSOPTIONcompsoc
  \usepackage[nocompress]{cite}
\else
  \usepackage{cite}
\fi
\usepackage{amsmath,amssymb,amsfonts}
\usepackage{algorithmic}
\usepackage{graphicx}
\usepackage{subfigure}
\usepackage{epstopdf}
\usepackage{caption}
\usepackage{cite}
\usepackage{url}
\usepackage{stfloats}

\begin{document}

\title{A Two-step Calibration Method for Unfocused Light Field Camera Based on Projection Model Analysis}

\author{Dongyang Jin,
        Saiping Zhang, 
        Xiao Huo,
        Wei Zhang,
        and Fuzheng Yang, 
}

\markboth{IEEE TRANSACTIONS ON PATTERN ANALYSIS AND MACHINE INTELLIGENCE,~Vol.~, No.~,}%
{Shell \MakeLowercase{\textit{et al.}}: Bare Advanced Demo of IEEEtran.cls for IEEE Computer Society Journals}

\IEEEtitleabstractindextext{%
\begin{abstract}
Accurately calibrating light field camera is essential to its applications. Rapid progress has been made in this area in the past decades. In this paper, detailed analysis was first performed towards the state of the art projection models for calibration which were further interpreted in three representations, including the correspondence between rays and pixels, 3D physical points and pixels and between 3D physical points and 3D signal structure of the captured light field. Based on the analysis, parameters in the projection model were grouped into direction parameter set and depth parameter set. A two-step calibration method was then proposed with each step dealing with each set of parameters. The proposed method is able to reuse traditional camera calibration methods for the direction parameter set. A simply raw image-based calibration of depth parameter set was further proposed. Systematic validations were conducted to evaluate the performance of the proposed calibration method. Experimental results show that the accuracy and robustness of the proposed method outperforms its counterparts under various benchmark criteria.

\end{abstract}

}

\maketitle

\IEEEdisplaynontitleabstractindextext

\IEEEpeerreviewmaketitle

\ifCLASSOPTIONcompsoc
\IEEEraisesectionheading{\section{Introduction}\label{sec:introduction}}

\begin{figure*}[tb]
\centering
\includegraphics[width=150mm]
{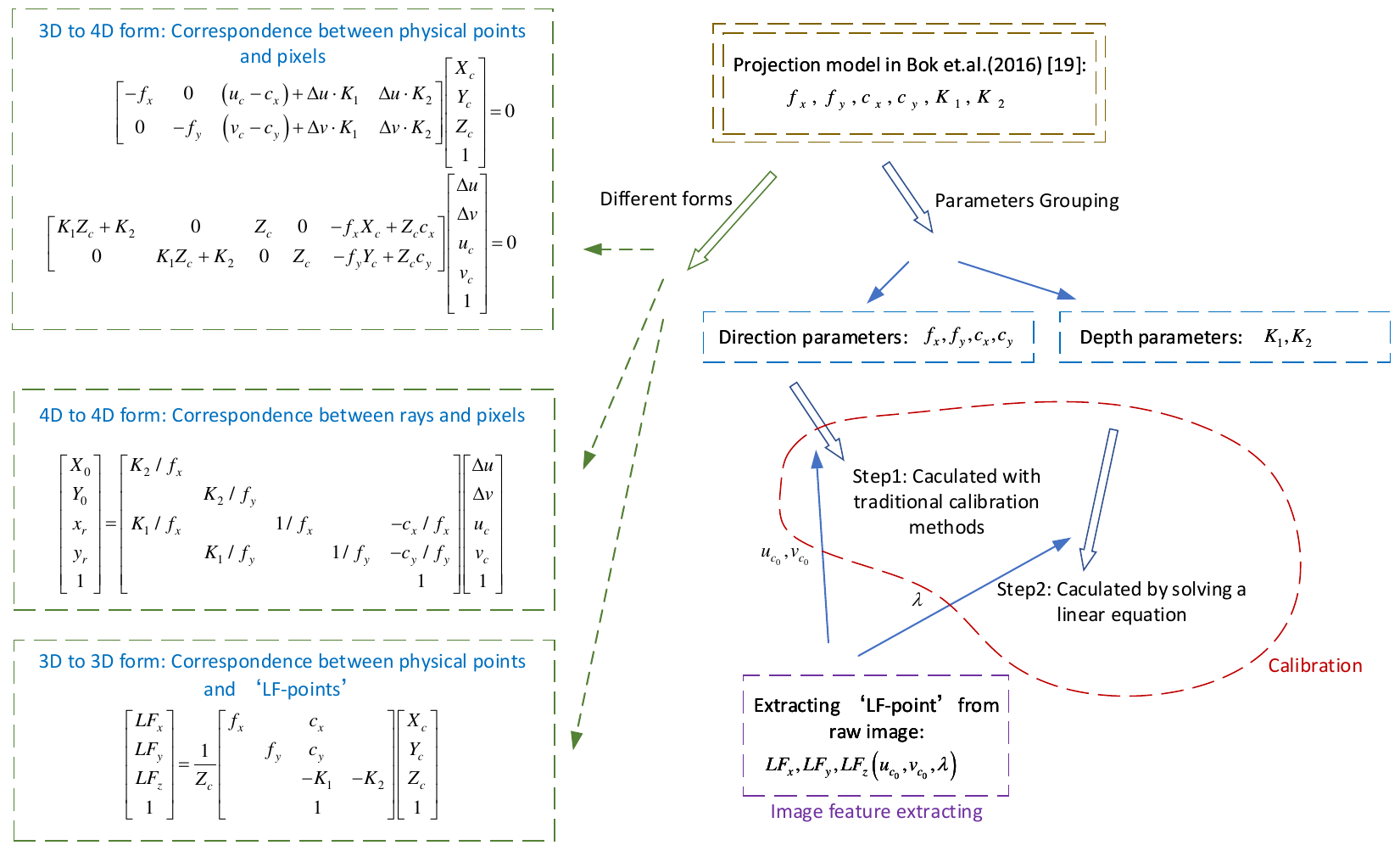}
\caption{Outline of this paper.}
\label{fig:Total}
\end{figure*}

\IEEEPARstart{R}{ecent} years have witnessed tremendous development in light field technologies. Light field cameras, devices to collect light rays on the image sensor in a single photographic exposure in both spatial and angular dimensions, were designed to realize the light field technique in various engineering applications. Due to the capability of recording angular information of the light rays from the external world, light field cameras have boosted the development of the computational photography and computer vision in a wide variety of applications, including refocusing~\cite{ng2005fourier}, depth estimation~\cite{jeon2015accurate},~\cite{tao2013depth},~\cite{wanner2012globally}, Synthetic Aperture Imaging (SAI)~\cite{vaish2006reconstructing},~\cite{yang2014all}, and Visual Simultaneous Localization and Mapping (Visual SLAM)~\cite{dong2013plenoptic},~\cite{zeller2017calibration}.

Many types of light field cameras were proposed~\cite{wilburn2005high}~\cite{veeraraghavan2007dappled}~\cite{liang2008programmable}~\cite{marwah2013compressive}~\cite{taguchi2010axial} in the past decades. Among those, micro-lens array (MLA) based light field cameras were the most popularized due to their advantages in low manufacturing expenditure and high-quality generation of light field images. State of the art MLA-based light field cameras include unfocused light field camera designed by Ng. et al. ~\cite{nglight} and focused light field camera designed by Georgiev and Lumsdaine ~\cite{lumsdaine2009focused}. The two types differ in the role MLA plays. The former is splitting light and the latter focusing the image behind the main lens. Well-known commercial MLA-based light field cameras developed on the basis of~\cite{nglight} and~\cite{lumsdaine2009focused} are Lytro~\cite{Lytroorg} and Raytrix~\cite{Raytrixorg}.

To achieve an optimal performance in various applications, especially in 3D reconstruction and visual SLAM, it is of fundamental important to accurately calibrate light field cameras. In recent years, various research has been conducted to calibrate unfocused light field cameras (e.g. ~\cite{dansereau2013decoding},~\cite{bok2016geometric},~\cite{noury2017light},~\cite{zhang2018generic},~\cite{o2018calibrating}). Different projection models were proposed to represent the unfocused light field camera. Different image features were used to represent the correspondence between pixels on the camera sensor and 3D physical points in real scenes. Additionally, different computing methods were proposed to estimate the intrinsic parameters of cameras.

Though rapid progress has been made, some research questions remain to be answered. For example, there is a lack of a uniform form for expressing the camera's projection model. It is also under discussion how to effectively apply the raw image in the calibration pipeline. State of the art pioneering work in~\cite{bok2016geometric} first used raw image in calibrating unfocused light field camera. Moreover, line features instead of point features were used for calibration. Experimental results demonstrate the effectiveness of this method. Therefore, in this paper, we proposed a calibration method on the basis of the work in~\cite{bok2016geometric} with improvements in three aspects, including a new expression of the projection model to accommodate different applications, a classification of the parameters in the model based on their meanings and a simple but efficient two-step parameter solving in the calibration pipeline.

Our contributions are as follow:

1)	We interpret the projection model in~\cite{bok2016geometric} in three forms, including the one describing the relationship between rays and pixels, the one describing the relationship between 3D physical points and pixels and the one describing the relationship between 3D physical point and 3D signal structure of the captured light field. The three proposed expression forms facilitate the usage of light field cameras in different applications.

2)	By analyzing the meanings and functionalities of all the parameters of the projection model, a parameter grouping strategy is proposed to divide the intrinsic parameters into direction parameter set and depth parameter set. These two parameter sets are also calculated separately during the calibration process. By doing so, the calibration methods for traditional cameras can be reused for calibrating the light field camera. The whole calibration pipeline is also simplified while the calibration accuracy is improved.

3)	We propose a new image feature called 'LF-point' which is a description of the signal structure of the light field data. Based on our previous work~\cite{previousCorner}, we also propose an extraction method of the 'LF-point' by making full use of the raw image instead of detecting corners in sub-aperture images. The proposed method can also export disparity information of each corner without extracting Epipolar-Plane Images (EPI), which facilitate the calibration of depth related intrinsic parameters.

\section{Related Works}

Numerous efforts have been made to solve the calibration of light field cameras ~\cite{dansereau2013decoding},~\cite{bok2016geometric},~\cite{noury2017light},~\cite{zhang2018generic},~\cite{o2018calibrating},~\cite{zhou2019two},~\cite{duan2019new},~\cite{heinze2016automated},~\cite{nousias2017corner}. A detailed analysis towards these methods is given in this section focusing on three aspects in the calibration pipeline, namely the camera representation, parameter determination and image feature detection.

\textbf{Camera representation}

Generally speaking, the light field camera representations in the literature can be divided into two categories. The first category represents the camera with its inherent physical parameters. Research in~\cite{johannsencalibration} aims at calibrating raytrix camera which is an enhanced version of the focused light field camera~\cite{lumsdaine2009focused}. Several parameters, such as the focal length $f$, the focus distance $h$ and the distance between MLA and sensor $b$ were used to represent the light field camera. Likewise, research in~\cite{zhou2019two} studied the usage of two sets of physical parameters, main lens related set and MLA related set, to represent the unfocused light field camera.

Another method for representing light field cameras is to use the so-called projection model which is in a more compact form with fewer parameters. Most of the methods proposed recently fall into this category. Dansereau et al.~\cite{dansereau2013decoding} was the first to deliver an end-to-end geometric calibration method for unfocused light field camera. A 12-parameter homogeneous matrix was proposed to represent the projection model which connects every pixel on the camera sensor and the corresponding rays coming into the camera. Zhang et al.~\cite{zhang2018generic} proposed a multi-projection-center (MPC) model with six intrinsic parameters to characterize both unfocused and focused light field cameras. The projection model efficiently connects 3D geometry of the real world to the recorded 4D light field.

One state of the art is the area done by Bok et al.~\cite{bok2016geometric} who proposed a compact projection model with only six parameters. In~\cite{bok2016geometric}, this projection model is used to construct the correspondence between 'line features' on the raw image and checkerboard outside the camera. Due to the satisfactory performance of~\cite{bok2016geometric}, several studies have been carried out as the extension of this one~\cite{bok2016geometric}. For example, OBrien et al.~\cite{o2018calibrating} derived a projection model connecting 3D physical points to a 3-dimensional 'plenoptic disc feature' with a one-to-one mapping. Similarly, Sotiris Nousias~\cite{nousias2017corner} also used the projection model in~\cite{bok2016geometric} for determining the initial solution of intrinsic and extrinsic parameters. The correspondence used in~\cite{nousias2017corner} is 3D physical points to 2D corners which is different from~\cite{bok2016geometric} where correspondence between 3D physical points to 2D line features is used for calibration.

Based on the analysis, the expression form of the projection model can be various although the physical structure of the unfocused light field camera is fixed. In this paper, different expression forms of the projection model in~\cite{bok2016geometric} is interpreted and proposed to facilitate its usage in various applications.

\textbf{Parameters Solution}

Once the projection model is determined, a key step in the calibration process is to solve the parameters in the projection model. Many calibration methods including the methods introduced above~\cite{bok2016geometric, o2018calibrating, nousias2017corner} solve all the parameters in the projection model jointly. Specifically, in these papers, initial solutions of all parameters are calculated together by linear regression followed by a joint optimization.

Additionally, some methods determine the parameters in groups based on their different meanings. Research in~\cite{strobl2016stepwise} divided parameters into two groups including the ones related to the brightness image and the ones related to the depth image. Parameters related to the brightness image were first determined and parameters related to the depth images were solved subsequently. Moreover, these two groups of parameters were also optimized separately. By leveraging existing pinhole camera calibration methods in solving the first group of parameters, the complexity of the entire calibration pipeline is reduced while higher accuracy of the calibration result can be achieved. For unfocused light field camera, research in~\cite{zhou2019two} also divided the parameters in the projection model into two groups with one corresponding to the main lens and the rest corresponding to the MLA. Experimental results also showed better results. In this paper, we also choose to divide the parameters of the projection model into two groups and deal with them separately in the calibration pipeline.

\textbf{Image feature detection}

Image feature detection (e.g. corner detection) is another important module in the entire calibration pipeline. The extraction of image features with high precision is essential to create accurate correspondence between pixels on the camera sensor and points on the checkerboard used for calibration. However, the precision of the detected corner locations is a bottleneck for the calibration of unfocused light field camera. Most calibration methods for unfocused light field camera are based on sub-aperture images. In methods~\cite{dansereau2013decoding} and~\cite{zhang2018generic}, sub-aperture images were first extracted from the raw images. Then corner locations in sub-aperture images were detected as point features for calibration. In~\cite{o2018calibrating} the 'plenoptic disc feature' was calculated from corner locations in sub-aperture image for calibrating unfocused light field camera. In the calibration pipeline in~\cite{zhou2019two}, the disparity information was calculated with a line fitting method which uses corner locations in all sub-aperture images as input. However, the resolution of sub-aperture image is too low compared with raw image. Taking Lytro Illum camera as an example, the resolution of its raw image is $7728\times 5368$ whilst the resolution of a sub-aperture image extracted from the raw image is only $625\times 434$. The loss of information during the sub-aperture image extraction process can significantly reduce the calibration accuracy of light field camera.

In our previous work~\cite{previousCorner}, by modifying the work in~\cite{bok2016geometric}, a method to directly obtain the corner locations on raw image was proposed. Based on this work, a new image feature called 'LF-point' representing the signal structure of the light field data is defined and extracted from the raw image in this paper. The new feature is able to provide high accuracy corner locations and also provide disparity information for every corner.

\section{Analysis of the projection model}

The terms that are defined and commonly used in this paper are listed in advance.

\begin{enumerate}
\item 3D physical point: a scene point outside the light field camera.
\item 3D corner: a corner on the checkerboard.

\item 2D projected point: the projected point of a 3D physical point in the raw image or sub-aperture image.

\item 2D corner: the projected point of a 3D corner in the raw image or sub-aperture image.
\end{enumerate}

Light field camera plays different roles in its different applications. For example, in view synthesis and digital refocus, it aims at obtaining rays outside the camera from the pixels on its sensor. In depth estimation and 3D reconstruction, it aims at calculating depth of the 3D physical points from the disparity of the recorded light field. To make the light field camera accommodate different applications, its projection model should have different expression forms constructing different correspondence in these applications.

Many projection models were created to bridge the relationship between pixels on the sensor and points in the real world in the past decades. One representative is the projection model proposed in~\cite{bok2016geometric}. In this section, this model is analyzed in detail with three matrix forms with each of them corresponding to the 3D (physical points) to 4D (pixels), 4D (rays) to 4D (pixels) and 3D (physical points) to 3D (LF points). The first two forms are derived based on the work in~\cite{bok2016geometric} with the purposes of clarifying the underlying physical meaning as well as facilitating the usage of the projection model. More importantly, the third form is derived based on the above two forms to express the correspondence between 3D physical points and the 'LF-point'. All the meanings of the parameters in the projection model of~\cite{bok2016geometric} and the roles they played are also analyzed.

\subsection{A 3D to 4D form: Correspondence between physical points and pixels}
As introduced in~\cite{bok2016geometric}, the correspondence between 3D physical points to pixels is expressed as

\begin{equation}\label{orgPR}
\left[ \begin{matrix}
   \Delta u  \\
   \Delta v  \\
\end{matrix} \right]=\frac{1}{{{K}_{1}}{{Z}_{c}}+{{K}_{2}}}\left[ \begin{matrix}
   {{f}_{x}}{{X}_{c}}-{{Z}_{c}}\left( {{u}_{c}}-{{c}_{x}} \right)  \\
   {{f}_{y}}{{Y}_{c}}-{{Z}_{c}}\left( {{v}_{c}}-{{c}_{y}} \right)  \\
\end{matrix} \right]
\end{equation}
It shows the correspondence between a 3D physical point $P$ outside the camera and a pixel $p$ on the camera sensor. In this equation, the coordinates of the 3D physical point $P$ in the camera coordinate system is denoted by $\left( X_c, Y_c, Z_c \right)$. The location of the pixel $p$ on the sensor is determined by the four parameters $\left( u_c, v_c, \Delta u, \Delta v \right)$, where $u_c, v_c$ represents the center of a micro-lens image, and $\Delta u, \Delta v$ denotes the displacement of this pixel $p$ from $u_c, v_c$(~\cite{bok2016geometric}), as shown in Fig.~\ref{fig:ParaMicro}.
\begin{figure}[bth]
\centering
\includegraphics[width=70mm]
{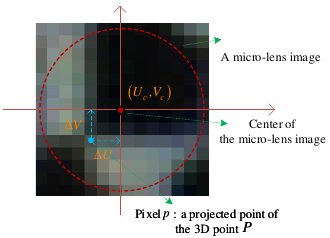}
\caption{Parameters to locate a pixel on the raw image.}
\label{fig:ParaMicro}
\end{figure}
The remaining six symbols $f_x,f_y,c_x,c_y,K_1,K_2$ in this equation are intrinsic parameters of the light field camera. In other words, the projection model of~\cite{bok2016geometric} is composed with these six parameters. To clearly illustrate the correspondence, equation~\eqref{orgPR} is re-formed with the following two matrix expressions:
\begin{equation}\label{PR2}
\left[ \begin{matrix}
   -{{f}_{x}} & 0 & \left( {{u}_{c}}-{{c}_{x}} \right)+\Delta u\cdot {{K}_{1}} & \Delta u\cdot {{K}_{2}}  \\
   0 & -{{f}_{y}} & \left( {{v}_{c}}-{{c}_{y}} \right)+\Delta v\cdot {{K}_{1}} & \Delta v\cdot {{K}_{2}}  \\
\end{matrix} \right]\left[ \begin{matrix}
   {{X}_{c}}  \\
   {{Y}_{c}}  \\
   {{Z}_{c}}  \\
   1  \\
\end{matrix} \right]=0
\end{equation}
and
\begin{equation}\label{PR3}
\begin{aligned}
& \left[ \begin{matrix}
   {{K}_{1}}{{Z}_{c}}+{{K}_{2}} & 0 & {{Z}_{c}} & 0 & -{{f}_{x}}{{X}_{c}}+{{Z}_{c}}{{c}_{x}}  \\
   0 & {{K}_{1}}{{Z}_{c}}+{{K}_{2}} & 0 & {{Z}_{c}} & -{{f}_{y}}{{Y}_{c}}+{{Z}_{c}}{{c}_{y}}  \\
\end{matrix} \right]\cdot \\
& \left[ \begin{matrix}
   \Delta u  \\
   \Delta v  \\
   {{u}_{c}}  \\
   {{v}_{c}}  \\
   1  \\
\end{matrix} \right]=0.
\end{aligned}
\end{equation}
In Eq.~\eqref{PR2}, the 3D physical point can be easily calculated using all its corresponding projected pixels in the raw image. Equation~\eqref{PR3} reveals that the set of all rays intersecting a single 3D physical point forms a linear 2D subspace in the homogeneous 4D light field coordinate domain~\cite{johannsen2015on}, and it is easy to find the corresponding area of a 3D physical point in the recorded light field by this form. It is convenient to use these two forms in light field camera intrinsic calibration (e.g. the method~\cite{bok2016geometric} itself), and structure from motion which often requires to construct correspondence between 3D physical points and the recorded light field.

\subsection{A 4D to 4D form: Correspondence between rays and pixels}

Using the six intrinsic parameters $f_x,f_y,c_x,c_y,K_1,K_2$ of light field cameras, research in~\cite{bok2016geometric} expressed the correspondence between rays outside the camera and pixels on the sensor as:
\begin{equation}\label{orgRP1}
\left[ \begin{matrix}
   {{X}_{0}}  \\
   {{Y}_{0}}  \\
\end{matrix} \right]={{K}_{2}}\left[ \begin{matrix}
   \Delta u/{{f}_{x}}  \\
   \Delta v/{{f}_{y}}  \\
\end{matrix} \right]
\end{equation}

\begin{equation}\label{orgRP2}
\left[ \begin{matrix}
   {{x}_{r}}  \\
   {{y}_{r}}  \\
\end{matrix} \right]={{K}_{1}}\left[ \begin{matrix}
   \Delta u/{{f}_{x}}  \\
   \Delta v/{{f}_{y}}  \\
\end{matrix} \right]+\left[ \begin{matrix}
   \left( {{u}_{c}}-{{c}_{x}} \right)/{{f}_{x}}  \\
   \left( {{v}_{c}}-{{c}_{y}} \right)/{{f}_{y}}  \\
\end{matrix} \right]
\end{equation}
where $X_0, Y_0$ denotes the intersection of a ray with the main-lens plane, and $x_r, y_r$ denotes the direction of this ray. Consider (4) and (5), equation (6) is derived with the left side representing a ray outside the camera and the right side being the homogeneous coordinates $\left(\Delta u, \Delta v, u_c, v_c, 1\right)$ locating a pixel on the sensor. More specifically, equation~\eqref{RP3} represents the relationship between rays and pixels. This form facilitates the view synthesis and light field rendering between light field cameras at different locations.
\begin{equation}\label{RP3}
\left[ \begin{matrix}
   {{X}_{0}}  \\
   {{Y}_{0}}  \\
   {{x}_{r}}  \\
   {{y}_{r}}  \\
   1  \\
\end{matrix} \right]=\left[ \begin{matrix}
   {{K}_{2}}/{{f}_{x}} & {} & {} & {} & {}  \\
   {} & {{K}_{2}}/{{f}_{y}} & {} & {} & {}  \\
   {{K}_{1}}/{{f}_{x}} & {} & 1/{{f}_{x}} & {} & -{{c}_{x}}/{{f}_{x}}  \\
   {} & {{K}_{1}}/{{f}_{y}} & {} & 1/{{f}_{y}} & -{{c}_{y}}/{{f}_{y}}  \\
   {} & {} & {} & {} & 1  \\
\end{matrix} \right]\left[ \begin{matrix}
   \Delta u  \\
   \Delta v  \\
   {{u}_{c}}  \\
   {{v}_{c}}  \\
   1  \\
\end{matrix} \right]
\end{equation}

\subsection{A 3D to 3D form: Correspondence between physical points and LF-points}
\label{sec:3Dto3DBig}

Previous subsections describe two forms of the projection model. In the application of depth estimation using light field camera, however, these two forms fail to address the relationship between disparity and depth in a straightforward manner which is commonly used in depth estimation~\cite{wanner2014variational}~\cite{wanner2012globally}. With this regard, another expression form is derived in this section to correlate disparity and depth by constructing a 3D to 3D correspondence between inside and outside of the camera. This form can be easily applied for depth estimation and 3D reconstruction. Before detailing the expression form, an explanation towards the meaning of the six parameters in the projection model is given.

\subsubsection{Meanings of  $f_x, f_y, c_x, c_y$}\label{sec:fxfycxcy}

In~\cite{bok2016geometric}, the four parameters $f_x, f_y, c_x, c_y$ are defined as the bridge between the normalized coordinate and the image coordinate. In other words, they are the intrinsic parameters of the pinhole camera whose optical center is set at the center of the main lens. In equation \eqref{RP3}, in the condition that only the rays passing through the center of the main lens are considered (i.e., $X_0, Y_0$ are zero), equation~\eqref{RP3} can be simplified as
\begin{equation}\label{u0v0_k0l0}
\left[ \begin{matrix}
   {{x}_{r}}  \\
   {{y}_{r}}  \\
   1  \\
\end{matrix} \right]=\left[ \begin{matrix}
   1/{{f}_{x}} & {} & -{{c}_{x}}/{{f}_{x}}  \\
   {} & 1/{{f}_{y}} & -{{c}_{y}}/{{f}_{y}}  \\
   {} & {} & 1  \\
\end{matrix} \right]\left[ \begin{matrix}
   {{u}_{{{c}_{0}}}}  \\
   {{v}_{{{c}_{0}}}}  \\
   1  \\
\end{matrix} \right],
\end{equation}
where $u_{c_0}, v_{c_0}$ denote the $u_c, v_c$ coordinates when $\Delta u$ and $\Delta v$ equal zero. It can be further reformed as \begin{equation}\label{u0v0_P}
\left[ \begin{matrix}
   {{u}_{{{c}_{0}}}}  \\
   {{v}_{{{c}_{0}}}}  \\
   1  \\
\end{matrix} \right]=\left[ \begin{matrix}
   {{f}_{x}} & {} & {{c}_{x}}  \\
   {} & {{f}_{y}} & {{c}_{y}}  \\
   {} & {} & 1  \\
\end{matrix} \right]\left[ \begin{matrix}
   {{x}_{r}}  \\
   {{y}_{r}}  \\
   1  \\
\end{matrix} \right]=\frac{1}{{{Z}_{c}}}\left[ \begin{matrix}
   {{f}_{x}} & {} & {{c}_{x}}  \\
   {} & {{f}_{y}} & {{c}_{y}}  \\
   {} & {} & 1  \\
\end{matrix} \right]\left[ \begin{matrix}
   {{X}_{c}}  \\
   {{Y}_{c}}  \\
   {{Z}_{c}}  \\
\end{matrix} \right].
\end{equation}

From~\eqref{u0v0_P}, it can be found that $u_{c_0}, v_{c_0}$ are the 2D projected point of the 3D physical point $X_c,Y_c,Z_c$ in this pinhole camera. Furthermore, $f_x$ and $f_y$ represent the distance between the sensor plane and the main-lens plane measured in pixels, along $X$ and $Y$ axis respectively, as shown in Fig. \ref{fig:Para3D}. $c_x$ and $c_y$ are the offset of the origin of the image coordinate system from the principal point also measured in pixels.

Moreover, by setting the optical center of a sub-aperture~\cite{nghangheld} at the center of the main lens (i.e., denoted as center-view sub-aperture), $f_x, f_y, c_x, c_y$ can be seen as the intrinsic parameters of 'center-view sub-aperture' and $u_{c_0}, v_{c_0}$ can be treated as the 2D projected point of this 3D physical point $X_c, Y_c, Z_c$ in center-view sub-aperture image, as shown in Fig.~\ref{fig:Para3D}), which is measured in pixels of the raw image.

\begin{figure}[bth]
\centering
\includegraphics[width=80mm]
{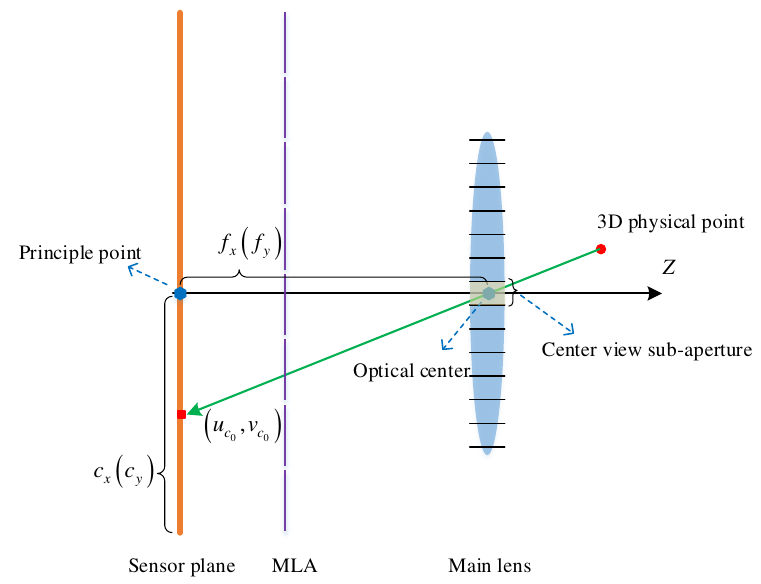}
\caption{2D projected point of a 3D physical point through center-view sub-aperture.}
\label{fig:Para3D}
\end{figure}

\subsubsection{Functionality of $K_1$ and $K_2$}
Suppose $p_1 = \left(\Delta u_1, \Delta v_1, u_{c_1}, v_{c_1}, 1 \right)$ and $p_2 = \left(\Delta u_2, \Delta v_2, u_{c_2}, v_{c_2}, 1 \right)$ are two points on the raw image which correspond to one 3D physical point, then we define
\begin{equation}\label{disp_k}
\lambda =\frac{{{u}_{{{c}_{1}}}}-{{u}_{{{c}_{2}}}}}{\Delta {{u}_{1}}-\Delta {{u}_{2}}},
\end{equation}
and
\begin{equation}\label{disp_l}
\lambda =\frac{{{v}_{{{c}_{2}}}}-{{v}_{{{c}_{2}}}}}{\Delta {{v}_{1}}-\Delta {{v}_{2}}}.
\end{equation}

Take $p_1 $ as an example, a brief introduction of the relationship between sub-aperture images and the four parameters $\Delta u, \Delta v, u_c, v_c$ is given. First, as defined in~\cite{bok2016geometric}, a sub-aperture image is a collection of camera rays that pass through a common point at the main-lens plane. Additionally, all rays corresponding to pixels with the same displacement $\Delta u, \Delta v$ pass through a common point at the main-lens plane~\cite{bok2016geometric}. Thus, the displacements $\Delta u_1, \Delta v_1$ of $p_1$ determine which sub-aperture image it corresponds to, and can be also considered as the index of this sub-aperture. Second, parameters $u_{c_1}, v_{c_1}$ represent the pixel coordinates of $p_1$ in its corresponding sub-aperture image. In fact, the pixel coordinate of $p_1$ is determined by the index of the micro-lens $p_1$ belongs to, and the index of this micro-lens in the MLA is proportional to the coordinate value of the center of its corresponding micro-lens image in the raw image, i.e. $u_c, v_c$. Technically, the index of a micro-lens in the MLA can be calculated by $\frac{1}{d_{MLA}}(u_c, v_c)$, where $d_{MLA}$ denotes the diameter of a micro-lens image measured in pixels. Thus, the pixel coordinates of $p_1$ in its corresponding sub-aperture image is proportional to the value of parameters $u_{c_1}, v_{c_1}$.

In conclusion, $\lambda$ in Eq.~\eqref{disp_k} and~\eqref{disp_l} is the ratio of the difference of the pixel coordinates of two projected points ($p_1$ and $p_2$) in two different sub-aperture images to the difference of the indices of these two sub-aperture images. That is to say, $\lambda$ is the disparity between the corresponding pixels of the same 3D physical point in two different sub-aperture images (along $u$ direction \eqref{disp_k} or along $v$ direction \eqref{disp_l}) after neglecting a scale factor $d_{MLA}$. Therefore, $\lambda$ is termed as disparity parameter.

Since $p_1 = \left(\Delta u_1, \Delta v_1, u_{c_1}, v_{c_1}, 1 \right)$ and $p_2 = \left(\Delta u_2, \Delta v_2, u_{c_2}, v_{c_2}, 1 \right)$ are two projected points corresponding to the same 3D physical point $X_c, Y_c, Z_c$, following relationship can be obtained using equation \eqref{orgPR} along $u$ direction,
\begin{equation}\label{orgPR1_orgPR2_i}
\begin{aligned}
  & \Delta {{u}_{1}}-\Delta {{u}_{2}}=\frac{1}{{{K}_{1}}{{Z}_{c}}+{{K}_{2}}}\left( {{f}_{x}}{{X}_{c}}-{{Z}_{c}}\left( {{u}_{{{c}_{1}}}}-{{c}_{x}} \right) \right) \\
 & -\frac{1}{{{K}_{1}}{{Z}_{c}}+{{K}_{2}}}\left( {{f}_{x}}{{X}_{c}}-{{Z}_{c}}\left( {{u}_{{{c}_{2}}}}-{{c}_{x}} \right) \right) ,\\
\end{aligned}
\end{equation}
which can be further simplified as
\begin{equation}\label{deltak_deltai_Z}
\frac{{{u}_{{{c}_{1}}}}-{{u}_{{{c}_{2}}}}}{\Delta {{u}_{1}}-\Delta {{u}_{2}}}=-{{K}_{1}}-\frac{{{K}_{2}}}{{{Z}_{c}}}
\end{equation}
by substituting Eq.~\eqref{disp_k} into \eqref{deltak_deltai_Z}, we get
\begin{equation}\label{disp_Z}
\lambda =-{{K}_{1}}-\frac{{{K}_{2}}}{{{Z}_{c}}}
\end{equation}

It should be noted that the same derivation can be obtained along $v$ direction. As shown in Eq.~\eqref{disp_Z}, parameters $K_1$ and $K_2$ are the bridge connecting the disparity parameter (i.e., lambda) and the depth of a 3D physical point in a real scene. In addition, $K_1$ and $K_2$ are independent of the $X$ and $Y$ coordinate components of the 3D physical point (i.e., $X_c$ and $Y_c$) and pixel coordinates of $p_1$ in its corresponding sub-aperture image (i.e., $u_c$ and $v_c$).

\subsubsection{Parameters Grouping}\label{sec:ParaGroup}

As introduced in subsection~\ref{sec:fxfycxcy}, $f_x, f_y, c_x, c_y$ are intrinsic parameters of the pinhole camera at the center-view sub-aperture to map the direction of a ray coming into the camera($x_r, y_r$) to a point in center view sub-aperture image ($u_{c_0},v_{c_0}$). They have no relationship to the exact depth this ray emits from. Additionally, $K_1$ and $K_2$ map the depth of a 3D physical point to a disparity parameter $\lambda$ with no indication on the relationship with the direction of rays this 3D physical point emits. In this regards, these six parameters of the projection model of~\cite{bok2016geometric} can be divided into two groups, namely the direction parameter set composed with $ f_x, f_y, c_x, c_y $ and the depth parameter set composed with $K_1, K_2$. These six parameters will also be determined by groups in following calibration procedure as described in section~\ref{sec:exp}.

\subsubsection{3Dto3D form}

\begin{figure}[bth]
\centering
\includegraphics[width=85mm]
{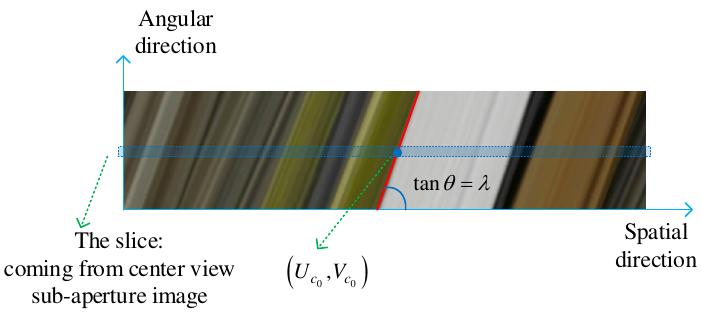}
\caption{The struture of the Epipolar Plane Image.}
\label{fig:EPI}
\end{figure}

Considering Eq. \eqref{disp_Z} and Eq. \eqref{u0v0_P}, the following can be obtained
\begin{equation}\label{3Dto3D}
\left[ \begin{matrix}
   {{u}_{{{c}_{0}}}}  \\
   {{v}_{{{c}_{0}}}}  \\
   \lambda   \\
   1  \\
\end{matrix} \right]=\frac{1}{{{Z}_{c}}}\left[ \begin{matrix}
   {{f}_{x}} & {} & {{c}_{x}} & {}  \\
   {} & {{f}_{y}} & {{c}_{y}} & {}  \\
   {} & {} & -{{K}_{1}} & -{{K}_{2}}  \\
   {} & {} & 1 & {}  \\
\end{matrix} \right]\left[ \begin{matrix}
   {{X}_{c}}  \\
   {{Y}_{c}}  \\
   {{Z}_{c}}  \\
   1  \\
\end{matrix} \right],
\end{equation}
It transforms a 3D physical point in the real scene to a triplet $\left( u_{c_0}, v_{c_0}, \lambda\right)$. It can be seen as a complete description of the structure of the captured 4D light field of this 3D physical point ($X_c, Y_c, Z_c$) with following reasons. First, all the rays that a 3D physical point emits forms a 2D affine subset in the space of 4D light field~\cite{gu1997polyhedral}, and the 2D affine subset intersect at a straight line with every Epipolar-Plane Image (EPI~\cite{bolles1987epipolarp}) of the 4D light field ~\cite{gortler1996lumigraph}. Furthermore, the location and the slope of the straight line in every EPI can be determined by the triplet $\left( u_{c_0}, v_{c_0}, \lambda\right)$. As shown in Fig.~\ref{fig:EPI}, the slope of the straight line is the disparity~\cite{gortler1996lumigraph}. The intersection of the straight line with the center-view sub-aperture image is ($u_{c_0}, v_{c_0}$). Moreover, this triplet $\left( u_{c_0}, v_{c_0}, \lambda\right)$ can be interpreted as a point coordinate with each component representing the $X$ ($u_{c_0}$), $Y$($v_{c_0}$) and depth information ($\lambda$) respectively. More specifically, the concept of virtual depth which is commonly used for focused light field ~\cite{johannsencalibration, PerwaSingle, heinze2016automated} is referred to explain its relationship between $\lambda$.

As shown in Fig.~\ref{fig:DispvirDp}, suppose $a$ denotes the distance between $P'$ (i.e., the image of the 3D physical point $P$ by main lens) and the sensor plane, and $b$ denotes the distance between the MLA plane and the sensor plane, then the virtual depth of $P'$ is defined as $v = a/b$(~\cite{johannsencalibration}).

\begin{figure}[bth]
\centering
\includegraphics[width=75mm]
{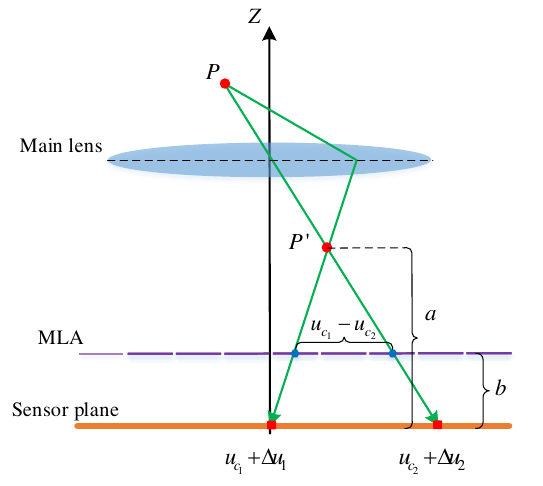}
\caption{Relationship between disparity parameter $\lambda$ and virtual depth.}
\label{fig:DispvirDp}
\end{figure}

From Fig.~\ref{fig:DispvirDp}, it is easy to obtain the following
\begin{equation}\label{similar_triangle1}
\frac{{{u}_{{{c}_{1}}}}-{{u}_{{{c}_{2}}}}}{{{\left( {{u}_{{{c}_{1}}}}+\Delta u \right)}_{1}}-\left( {{u}_{{{c}_{2}}}}+\Delta {{u}_{2}} \right)}=\frac{a-b}{a}
\end{equation}
which can be further simplified as
\begin{equation}\label{similar_triangle2}
\frac{{{u}_{{{c}_{1}}}}-{{u}_{{{c}_{2}}}}}{\left( {{u}_{{{c}_{1}}}}-{{u}_{{{c}_{2}}}} \right)+\left( \Delta {{u}_{1}}-\Delta {{u}_{2}} \right)}=\frac{a-b}{a},
\end{equation}
Therefore,
\begin{equation}\label{similar_triangle3}
\frac{1}{1+1/\lambda }=1-\frac{1}{v}
\end{equation}
i.e.
\begin{equation}\label{disp_vd}
\lambda =v-1
\end{equation}
It means that the disparity parameter and virtual depth only differ by one in numerical value indicating that $\lambda$ reflects the depth information of the virtual image $P'$. It therefore demonstrates that the triplet ($ u_{c_0}$, $ v_{c_0}$, and $\lambda$) reflects information along $X$ axis, $Y$ axis and $Z$ axis respectively. In this paper, we thus term the triplet as a 'light field point' (LF point) and rewrite the triplet as \[\left( L{{F}_{x}},L{{F}_{y}},L{{F}_{z}} \right)\]. Correspondingly, Eq.~\eqref{3Dto3D} is rewritten as
\begin{equation}\label{3Dto3D2}
\left[ \begin{matrix}
   L{{F}_{x}}  \\
   L{{F}_{y}}  \\
   L{{F}_{z}}  \\
   1  \\
\end{matrix} \right]=\frac{1}{{{Z}_{c}}}\left[ \begin{matrix}
   {{f}_{x}} & {} & {{c}_{x}} & {}  \\
   {} & {{f}_{y}} & {{c}_{y}} & {}  \\
   {} & {} & -{{K}_{1}} & -{{K}_{2}}  \\
   {} & {} & 1 & {}  \\
\end{matrix} \right]\left[ \begin{matrix}
   {{X}_{c}}  \\
   {{Y}_{c}}  \\
   {{Z}_{c}}  \\
   1  \\
\end{matrix} \right].
\end{equation}

It represents a projective transformation from a 3D physical point to a 'LF-point', which is the third form of the projection model of~\cite{bok2016geometric} proposed in this paper. We further term the signal space spanned by the three variables $\left(LF_x, LF_y, LF_z\right)$ as the light field structure space. Then, equation~\eqref{3Dto3D2} can be regarded as a projective transform from the 3D physical space to the light field structure space.

\section{LF-point extracting algorithm}\label{sec:extractingLFPt}

In this paper, the 3D to 3D form, i.e. the correspondence between physical points and 'LF-points' expressed in Eq.~\eqref{3Dto3D2} is used for calibration. In order to construct this correspondence, the 'LF-point' on the left side of Eq.~\eqref{3Dto3D2} is necessary to be extracted first from the recorded image. In this section, we will describe a method on calculating 'LF-point' ($\left(u_{c_0}, v_{c_0},\lambda\right)$) from the captured raw light field data for corners of the checkerboard.

As discussed in Sec. (\ref{sec:fxfycxcy}), $u_{c_0}, v_{c_0}$ is the 2D projected point of a 3D physical point in the center view sub-aperture image. Directly determining the value of $u_{c_0}, v_{c_0}$ by detecting the corner locations from the sub-aperture image is, however, inaccurate. This is due to that the accuracy of the corner detection can be affected to different extent depending on the sub-aperture image extraction methods used. Moreover, the lower resolution of the sub-aperture image (if compared with the raw image) also results in a reduction of corner detection accuracy.

As the raw image of the light field camera has the potential to provide higher accurate information than low resolution sub-aperture images~\cite{bishop2012the, Wanner2012Spatial}, in this paper, instead of using sub-aperture images, we propose a simple 'LF-point' extraction method directly from the raw image based on our previous work~\cite{previousCorner} . This work is able to directly obtain 2D corner locations in the raw image by calculating the intersection of two 3D line segments in the main-lens's image space and re-projecting the intersection to the raw image. The 2D corners corresponding to the same 3D corner is detected jointly. Relationships among these 2D corners can be preserved which improve the robustness and accuracy of the detected corner locations.

More specifically, in this 'LF-point' extraction method, the disparity parameter $\lambda$ is calculated simultaneously together with $u_{c_0}, v_{c_0}$. Since $\lambda$ is extracted from the raw image, the common procedure of extracting EPI images to obtain disparity can be avoided. Detailed steps are as follows.

\begin{figure}[bth]
\centering
\includegraphics[width=82mm]
{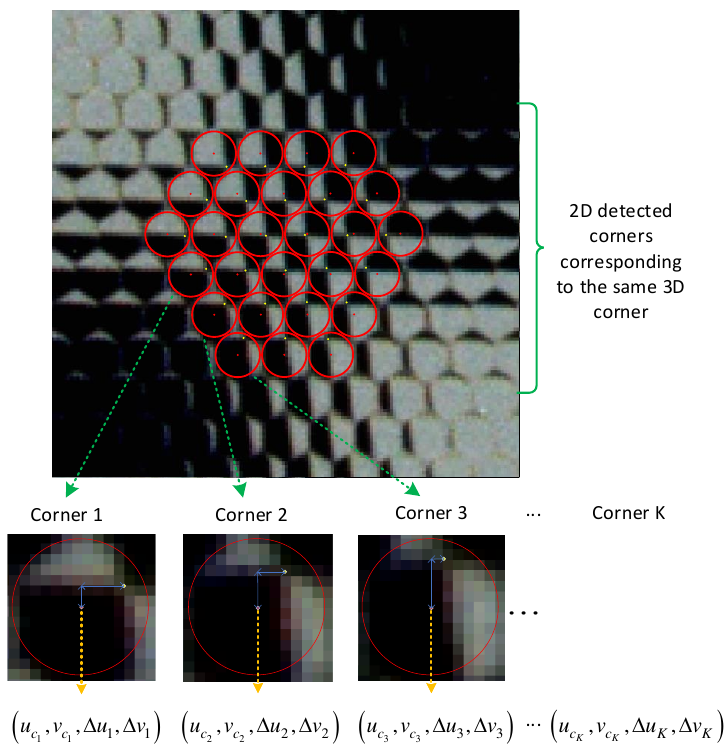}
\caption{2D detected corners corresponding to the same 3D corner.}
\label{fig:CornerMicrolensImg}
\end{figure}

For one 3D corner on the checkerboard, suppose $\left(u_{c_i}, v_{c_i}, \Delta u_i, \Delta v_i \right), i\in \left\{ 1\sim K \right\}$ are its $K$ 2D projected corners on the raw image detected by the method in~\cite{previousCorner}, as shown in Fig.~\ref{fig:CornerMicrolensImg}. With Eq. \eqref{disp_k} and \eqref{disp_l}, the relationship between each of them and $\left(u_{c_0}, v_{c_0}, \lambda\right)$ can be expressed as
\begin{equation}\label{disp_k0}
\lambda =\frac{{{u}_{{{c}_{0}}}}-{{u}_{{{c}_{i}}}}}{0-\Delta {{u}_{i}}}
\end{equation}
and
\begin{equation}\label{disp_l0}
\lambda =\frac{{{v}_{{{c}_{0}}}}-{{v}_{{{c}_{i}}}}}{0-\Delta {{v}_{i}}},
\end{equation}
which can also be expressed in a Matrix form as follow

\begin{equation}\label{stack_disp_k0l0_matrix}
\left[ \begin{matrix}
   1 & 0 & \Delta {{u}_{1}}  \\
   0 & 1 & \Delta {{v}_{1}}  \\
   1 & 0 & \Delta {{u}_{2}}  \\
   0 & 1 & \Delta {{v}_{2}}  \\
   \vdots  & \vdots  & \vdots   \\
   1 & 0 & \Delta {{u}_{K}}  \\
   0 & 1 & \Delta {{v}_{K}}  \\
\end{matrix} \right]\left[ \begin{matrix}
   {{u}_{{{c}_{0}}}}  \\
   {{v}_{{{c}_{0}}}}  \\
   \lambda   \\
\end{matrix} \right]=\left[ \begin{matrix}
   {{u}_{{{c}_{1}}}}  \\
   {{v}_{{{c}_{1}}}}  \\
   {{u}_{{{c}_{2}}}}  \\
   {{v}_{{{c}_{2}}}}  \\
   \vdots   \\
   {{u}_{{{c}_{K}}}}  \\
   {{v}_{{{c}_{K}}}}  \\
\end{matrix} \right].
\end{equation}
This equation can be solved using matrix decomposition algorithms, e.g., singular value decomposition (SVD) to obtain $u_{c_0}, v_{c_0}, \lambda$. Experimental results in section~\ref{sec:exp} show that this method can achieve higher corner detection accuracy than the solution using sub-aperture image.

\section{Calibration}
As referred in \cite{strobl2016stepwise} and \cite{zhou2019two}, the entire raw image is not always required to estimate each intrinsic parameter of the light field camera. Specifically, intrinsic parameters are divided into groups in these two papers, where the estimation of the main lens related parameters in [24] and the brightness image related parameters in [31] are performed without the dependence on the depth related information in the raw image. Similarly, the intrinsic parameters are also grouped in our method and we followed the idea in~\cite{strobl2016stepwise} and \cite{zhou2019two} by calibrating the intrinsic parameters separately using different information of the raw image. A very simple calibration method with two steps is proposed with the first step to determine the direction parameter set $ f_x, f_y, c_x, c_y $ by reusing the traditional calibration method~\cite{zhang1999flexible} and the second step to determine the depth parameter set $K_1, K_2$ by solving a simple linear equation. Experimental results in Sec. \ref{sec:exp} demonstrate that this simple method can achieve comparable or even better results than~\cite{bok2016geometric} and another state-of-art~\cite{dansereau2013decoding}.

\subsection{Direction parameter set}\label{sec:dirPara}
As referred in Sec.~\ref{sec:fxfycxcy}, $f_x, f_y, c_x, c_y$ are the intrinsic parameters of the pinhole camera at the center view sub-aperture. They are determined using the well-known calibration method in~\cite{zhang1999flexible} in following steps.

\subsubsection{Initial solution}

Equation \eqref{u0v0_P} is transformed to a more general form,
\begin{equation}\label{u0v0_P_general}
   ~\left[ \begin{matrix}
   {{u}_{{{c}_{0}}}}  \\
   {{v}_{{{c}_{0}}}}  \\
   1  \\
\end{matrix} \right]=\frac{1}{{{Z}_{c}}}\left[ \begin{matrix}
   {{f}_{x}} & {} & {{c}_{x}}  \\
   {} & {{f}_{y}} & {{c}_{y}}  \\
   {} & {} & 1  \\
\end{matrix} \right]\left[ \begin{matrix}
   R & T  \\
\end{matrix} \right]\left[ \begin{matrix}
   {{X}_{w}}  \\
   {{Y}_{w}}  \\
   {{Z}_{w}}  \\
   1  \\
\end{matrix} \right],
\end{equation}
where $\left[X_w, Y_w, Z_w\right]$ denote the world coordinate (checkerboard coordinate) of a 3D physical point. $R$ and $T$ denote the rigid transform from the world coordinate system to camera coordinate system in Eq.~\eqref{eq:wToc}.
\begin{equation}\label{eq:wToc}
\left[ \begin{matrix}
   {{X}_{c}}  \\
   {{Y}_{c}}  \\
   {{Z}_{c}}  \\
   1  \\
\end{matrix} \right]=\left[ \begin{matrix}
   R & T  \\
\end{matrix} \right]\left[ \begin{matrix}
   {{X}_{w}}  \\
   {{Y}_{w}}  \\
   {{Z}_{w}}  \\
   1  \\
\end{matrix} \right],
\end{equation}
where $R$ is a $3\times3$ rotation matrix and $T$ is a $3\times1$ translation matrix.

Considering $Z_w = 0$, by eliminating the $Z_w$ coordinate and the third column of the matrix $\left[R~T\right]$, Eq.~\eqref{u0v0_P_general} can be re-written as
\begin{equation}\label{u0v0_P_general_Zequals0}
\begin{aligned}
   \left[ \begin{matrix}
   {{u}_{{{c}_{0}}}}  \\
   {{v}_{{{c}_{0}}}}  \\
   1  \\
\end{matrix} \right]&=\frac{1}{{{Z}_{c}}}\left[ \begin{matrix}
   {{f}_{x}} & {} & {{c}_{x}}  \\
   {} & {{f}_{y}} & {{c}_{y}}  \\
   {} & {} & 1  \\
\end{matrix} \right]\left[ \begin{matrix}
   \begin{matrix}
   {{r}_{1}} & {{r}_{2}}  \\
\end{matrix} & T  \\
\end{matrix} \right]\left[ \begin{matrix}
   {{X}_{w}}  \\
   {{Y}_{w}}  \\
   1  \\
\end{matrix} \right]\\
&=\frac{1}{{{Z}_{c}}}H\left[ \begin{matrix}
   {{X}_{w}}  \\
   {{Y}_{w}}  \\
   1  \\
\end{matrix} \right]
\end{aligned},
\end{equation}
where $r_1$ and $r_2$ are the first and second column of the matrix $\left[R~T\right]$, and
\begin{equation}\label{H}
   H=\left[ \begin{matrix}
   {{f}_{x}} & {} & {{c}_{x}}  \\
   {} & {{f}_{y}} & {{c}_{y}}  \\
   {} & {} & 1  \\
\end{matrix} \right]\left[ \begin{matrix}
   \begin{matrix}
   {{r}_{1}} & {{r}_{2}}  \\
\end{matrix} & T  \\
\end{matrix} \right],
\end{equation}
which is a homography matrix defining a 2D projective transformation from $\left[X_w, Y_w, 1\right]$ to $\left[u_{c_0}, v_{c_0}, 1\right]$. By varying the relative location between the camera and the checkerboard, several $H$ can be obtained which include the same $f_x, f_y, c_x, c_y$ but different extrinsic parameters $\left[R~T\right]$. Using the calibration method detailed in~\cite{zhang1999flexible} which utilizes the orthogonality constraints of $r_1$ and $r_2$, the initial solution for $f_x, f_y, c_x, c_y$ and every $\left[R~T\right]$ is calculated.

\subsubsection{Nonlinear optimization}\label{sec:nonlinear}

To refine the initial solution, a nonlinear optimization process is performed subsequently. In the proposed method, the cost function for the optimization is defined as
\begin{equation}\label{min_proj_error}
\min \sum\limits_{i=1}^{N}{\sum\limits_{j=1}^{M}{{{\left\| {{p}^{i,j}}_{detect}-{{p}^{i,j}}_{proj}\left( {{f}_{x}},{{f}_{y}},{{c}_{x}},{{c}_{y}},R,T \right) \right\|}^{2}}}},
\end{equation}
where ${{p}^{i,j}}_{proj}$ is the 2D projected point in the center view sub-aperture image of the $j^{th}$ 3D corner in the $i^{th}$ photo calculated by equation~\eqref{u0v0_P_general_Zequals0}. ${{p}^{i,j}}_{detect}$ is the detected 2D corner in the center view image of the $i^{th}$ photo of the $j^{th}$ 3D corner calculated using the method described in Sec.~\ref{sec:extractingLFPt}. $N$ is the total number of all captured photos. $M$ is the total number of corners on the checkerboard. In the proposed method, the minimization of Eq.~\eqref{min_proj_error} is solved by the commonly used Levenberg-Marquardt algorithm.

\subsubsection{Distortion model}

Radial and tangential distortion~\cite{weng1992camera}~\cite{heikkila1997four} are both considered, i.e.
\begin{equation}\label{org_distortion}
   \begin{aligned}
  & {{x}_{dis}}=\hat{x}+{{x}_{rad}}+{{x}_{tan}} \\
 & {{y}_{dis}}=\hat{y}+{{y}_{rad}}+{{y}_{tan}}, \\
\end{aligned}
\end{equation}
where $\hat{x},\hat{y}$ and $x_{dis},y_{dis}$ denote the ideal (distortion-free) and real (distorted) normalized coordinates of $u_{c_0},v_{c_0}$, respectively. $x_{rad}, y_{rad}$ and $x_{tan}, y_{tan}$ denote the radial distortion component and tangential distortion component respectively, which are defined as follows,
\begin{equation}\label{rad_distortion}
\begin{aligned}
  & {{x}_{rad}}=\hat{x}\left( {{k}_{1}}\left( {{{\hat{x}}}^{2}}+{{{\hat{y}}}^{2}} \right)+{{k}_{2}}{{\left( {{{\hat{x}}}^{2}}+{{{\hat{y}}}^{2}} \right)}^{2}} \right) \\
 & {{y}_{rad}}=\hat{y}\left( {{k}_{1}}\left( {{{\hat{x}}}^{2}}+{{{\hat{y}}}^{2}} \right)+{{k}_{2}}{{\left( {{{\hat{x}}}^{2}}+{{{\hat{y}}}^{2}} \right)}^{2}} \right), \\
\end{aligned}
\end{equation}

\begin{equation}\label{tan_distortion}
\begin{aligned}
  & {{x}_{tan}}=2{{p}_{1}}\hat{x}\hat{y}+{{p}_{2}}\left( {{{\hat{x}}}^{2}}+{{{\hat{y}}}^{2}}+2{{{\hat{x}}}^{2}} \right) \\
 & {{y}_{tan}}=2{{p}_{2}}\hat{x}\hat{y}+{{p}_{1}}\left( {{{\hat{x}}}^{2}}+{{{\hat{y}}}^{2}}+2{{{\hat{y}}}^{2}} \right). \\
\end{aligned}
\end{equation}
The distortion components $k_1, k_2, p_1, p_2$ are also estimated in the nonlinear optimization procedure, and the cost function with distortion is
\begin{equation}\label{min_proj_error_distortion}
\begin{aligned}
& \min \sum\limits_{I=1}^{N}{\sum\limits_{j=1}^{M} }\\
& {{{\left\| {{p}^{i,j}}_{detect}-
{{p}^{i,j}}_{proj}\left( {{f}_{x}},{{f}_{y}},{{c}_{x}},{{c}_{y}},{{k}_{1}},{{k}_{2}},{{p}_{1}},{{p}_{2}},R,T \right) \right\|}^{2}}}.
\end{aligned}
\end{equation}
which is also solved by the commonly used Levenberg-Marquardt algorithm.

\subsection{Depth parameter set}\label{sec:depPara}

The correspondence between depth and disparity is used to estimate the depth parameter set $K_1, K_2$. For all the corners in all photos, considering the depth-disparity relationship denoted by~\eqref{stack_depth_disp_correspond}, the following can be obtained
\begin{equation}\label{stack_depth_disp_correspond}
\left[ \begin{matrix}
   -1 & -\frac{1}{{{Z}_{c}^{1}}}  \\
   -1 & -\frac{1}{{{Z}_{c}^{2}}}  \\
   \vdots  & \vdots   \\
   -1 & -\frac{1}{Z_{c}^{N\times M}}  \\
\end{matrix} \right]\left[ \begin{matrix}
   {{K}_{1}}  \\
   {{K}_{2}}  \\
\end{matrix} \right]=\left[ \begin{matrix}
   {{\lambda }_{1}}  \\
   {{\lambda }_{2}}  \\
   \vdots   \\
   {{\lambda }_{N\times M}}  \\
\end{matrix} \right],
\end{equation}
where $M$ is the total number of corners on the checkerboard, and $N$ is the total number of captured raw images. The final results of $K_1, K_2$ is then obtained by SVD algorithm.

Note that the solution of Eq.~\eqref{stack_depth_disp_correspond} can only be considered as an initial solution of $K_1, K_2$, i.e. a linear solution. It can be demonstrated in the following experimental results section that even without a nonlinear optimization, this linear solution has already achieved a comparable or even better results than~\cite{bok2016geometric} and ~\cite{dansereau2013decoding}.

\section{Experimental Results}
\label{sec:exp}

To verify the performance of the proposed method, multiple benchmarks are conducted on six datasets. Three of them are from~\cite{dansereau2013decoding}, which are captured by one Lytro camera with different focal settings and are denoted as D-A, D-B and D-E in this paper. The checkerboard size in these three datasets are $19\times19$ grid of 3.61 mm cells, $19\times19$ grid of 3.61 mm cells and $8\times6$ grid of 35 mm cells respectively. The fourth dataset is taken from~\cite{bok2016geometric} which is captured by a Lytro illum camera and is denoted as G-A in this paper. The checkerboard size used for G-A is $9\times6$ grid of 26.25mm cells. The remaining two datasets are captured by ourselves using a Lytro illum camera and are denoted as P-A and P-B. The checkerboard sizes are $8\times12$ grid of 29.92 mm cells and $8\times12$ grid of 22.25 mm cells respectively.

\subsection{Accuracy of corner location}

To demonstrate that the accuracy of corner location detected using the proposed LF-point calculation method (termed as raw image-based method) is higher than traditional methods which directly detecting corners in sub-aperture images (termed as sub-aperture image-based method), a comparison experiment in between is conducted. Since the actual locations of the 2D projected points corner are unknown for real scenes, directly measuring the detection error of $u_{c_0},v_{c_0}$ is unfeasible. Instead, we use the camera calibration results as the indicator of the corner detection accuracy since higher corner detection accuracy lead to higher accuracy of camera calibration.
The pipeline is as follows,

\begin{enumerate}
\item For each dataset, MATLAB toolbox LFToolbox V0.4~\cite{matlabTool} is used to generate raw image and center view sub-aperture image.
\item Use the method discussed in section~\ref{sec:extractingLFPt} to extract 'LF-point' for each corner in each raw image. Take the value of $u_{c_0}, v_{c_0}$ in each 'LF-point' as the result of 'raw image based method'.
\item Use the MATLAB function 'detectCheckerboardPoints' to detect the corners locations in each of the center view sub-aperture image. Take the detected corner locations as the result of the 'sub-aperture image-based method'.
\item For each group of the corner locations detected by the two methods mentioned above, the method in~\cite{zhang1999flexible} is used to calibrate the pinhole camera at the center view sub-aperture.
\item Calculate the 'point to point re-projection error' (abbreviated as $PP_{error}$) using equation \eqref{min_reproj_error_PP} for each version of the intrinsic parameters. The one with the lower $PP_{error}$ is considered to lead to higher corner detection accuracy.
\end{enumerate}

\begin{equation}\label{min_reproj_error_PP}
P{{P}_{error}}=\frac{1}{N\times M}\sum\limits_{N}{\sum\limits_{M}{\left\| {{P}_{actual}}-{{P}_{reprojected}} \right\|_{2}^{2}}}
\end{equation}

In equation \eqref{min_reproj_error_PP}, $P_{actual}$ denotes the locations of the actual 3D corner points on the checkboard and $P_{reprojected}$ denotes the re-projected 3D physical point on the checkerboard from detected 2D corners calculated by the inverse of equatuion~\eqref{u0v0_P_general_Zequals0}. $\left\| \cdot  \right\|_{2}^{2}$ denotes the Euclidean distance between two 3D physical points. $M$ is the total number of the corners on one checkerboard. $N$ is the total number of the captured photos. Note that '$PP_{error}$' is chosen since it is measured in millimeters and has no relationship to the units of the detected 2D corner points (the detected 2D corner points are measured in the size of a pixel on the raw image for the 'raw image based method' and in the size of a micro-lens for the 'sub-aperture image based method'.). Table~\ref{tabA} shows the $PP_{error}$ for the two methods. It demonstrates that the proposed 'raw image-based method' is more accurate than the 'sub-aperture image-based' method.

\begin{table}
  \centering
  \caption{PP errors for the Raw Image Based and Sub-Aperture Image Based methods.}
  \begin{tabular}{ccc}
\hline
 Datasets & 	Sub-Aperture Image Based &	Raw Image Based \\
    \hline
D-A &	0.0722 &	0.0723  \\
D-B &	0.0309 &	0.0294   \\
D-E &	0.1014 &	0.0939   \\
P-A &	0.0517 &	0.0183   \\
P-B &	0.0285 &	0.0146   \\

 \hline
 \end{tabular}
  \label{tabA}
\end{table}

\begin{table}
  \centering
  \caption{Point-to-point errors (mm)}
  \begin{tabular}{cccc}
\hline
 Datasets & 	Proposed &	Dansereau ~\cite{dansereau2013decoding} &	Bok ~\cite{bok2016geometric} \\
    \hline
    D-A &	0.0977 &	0.0984 &	0.2711 \\
D-B &	0.0411 &	0.0442 &	0.1525  \\
D-E &	0.1733 &	0.1462 &	0.5404  \\
P-A &	0.0186 &	0.0622 &	0.2076  \\
P-B &	0.0157 &	0.0398 &	0.1392  \\
G-A &	0.1018 &	- &	0.2349  \\
 \hline
 \end{tabular}
  \label{tab1}
\end{table}

\begin{table}
  \centering
  \caption{Point to ray errors (mm)}
  \begin{tabular}{ccccc}
    \hline
 Algorithm & 	Proposed &	Dansereau ~\cite{dansereau2013decoding} &	Bok ~\cite{bok2016geometric} \\
\hline
D-A &	0.0806 &	0.0815 &	0.1076  \\
D-B &	0.0389 &	0.0420 &	0.0714   \\
D-E &	0.1639 &	0.1344 &	0.454   \\
P-A &0.0179 &0.0598 &0.1972   \\
P-B &0.0149 &0.0381 &0.1330  \\
G-A &0.0895 &- &0.2066   \\
\hline
 \end{tabular}
  \label{tab2}
\end{table}

\begin{table}
  \centering
  \caption{Relative depth errors}
  \begin{tabular}{cccc}
    \hline

   Dataset &	Proposed	& Dansereau ~\cite{dansereau2013decoding}	& Bok ~\cite{bok2016geometric} \\
    \hline
 D-A & 1.85 $\%$    & 3.14$\%$		& 8.96$\%$ \\
 D-B &	1.75 $\%$	& 1.82 $\%$	& 9.44$\%$ \\
 D-E &	12.94 $\%$	& 25.29 $\%$	& 29.22$\%$ \\
 P-A &	1.94 $\%$	& 10.35 $\%$	& 12.71$\%$ \\
 P-B &	2.33$\%$	& 4.95 $\%$	& 13.52$\%$ \\
 G-A	& 3.33$\%$	& -	$\%$	& 17.98 $\%$  \\
 \hline
 \end{tabular}
  \label{tab3}
\end{table}

\subsection{Re-projection Error of the Calibration Results}

To comprehensively evaluate the proposed calibration method, three metrics are used to measure the calibration errors. Since the objective function used in the proposed method does not use these metrics, there is no inherent advantage for the proposed method when performing the comparison.

\begin{figure}[bth]
\centering
\includegraphics[width=80mm]
{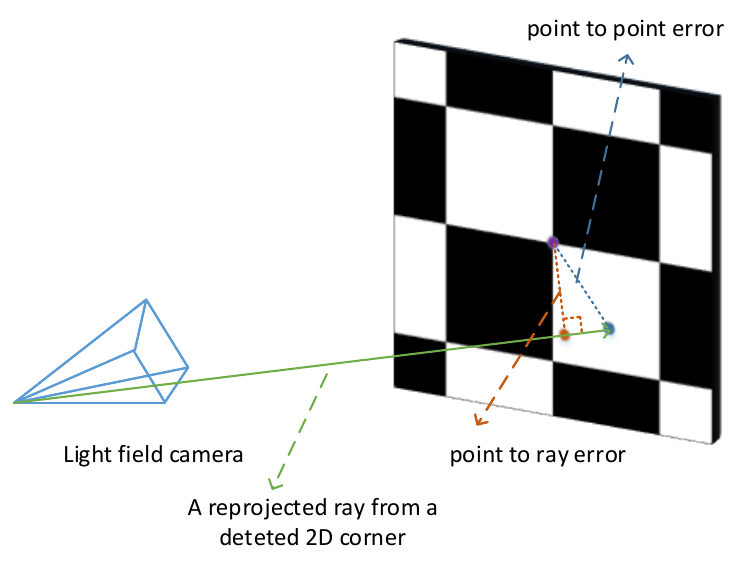}
\caption{The 'point to ray error' and 'point to point error'.}
\label{fig:PPPRerror}
\end{figure}

The first metric is the 'point to ray error' (denoted as PR error) which measures the distance from the actual 3D corner to the re-projected ray that a detected 2D corner point corresponds, as shown in Fig.~\ref{fig:PPPRerror}. The second metric is the 'point to point' error (denoted as PP error) which measures the distance between the actual 3D corner to the re-projected point of a detected 2D corner point on the checkerboard plane, as shown in Fig. \ref{fig:PPPRerror}. The third metric is the relative depth error (denoted as RDE) which measures the error between the actual 3D corner and the estimated 3D corner point along $Z$ axis, i.e. along depth direction. The calculation method is shown in equation \eqref{eq:relativeDp}, where $Z_{est}$ is the $Z$ coordinate of the estimated 3D corner point calculated by equation \eqref{disp_Z}, and $Z_{act}$ is the $Z$ coordinate of the actual 3D corner calculated by equation \eqref{eq:wToc}.

\begin{equation}
RDE=\left| \frac{{{Z}_{act}}-{{Z}_{est}}}{{{Z}_{est}}} \right|
\label{eq:relativeDp}\end{equation}

The PR error and PP error aims to measure the lateral errors whilst the RDE aims to measure the errors in depth direction.

The proposed method is compared against two state of the arts in~\cite{bok2016geometric} and~\cite{dansereau2013decoding}. Table~\ref{tab1}, Table~\ref{tab2} and Table~\ref{tab3} shows the results corresponding to PR error, PP error and RDE respectively.

\begin{figure}[bth]
\centering
\includegraphics[width=80mm]
{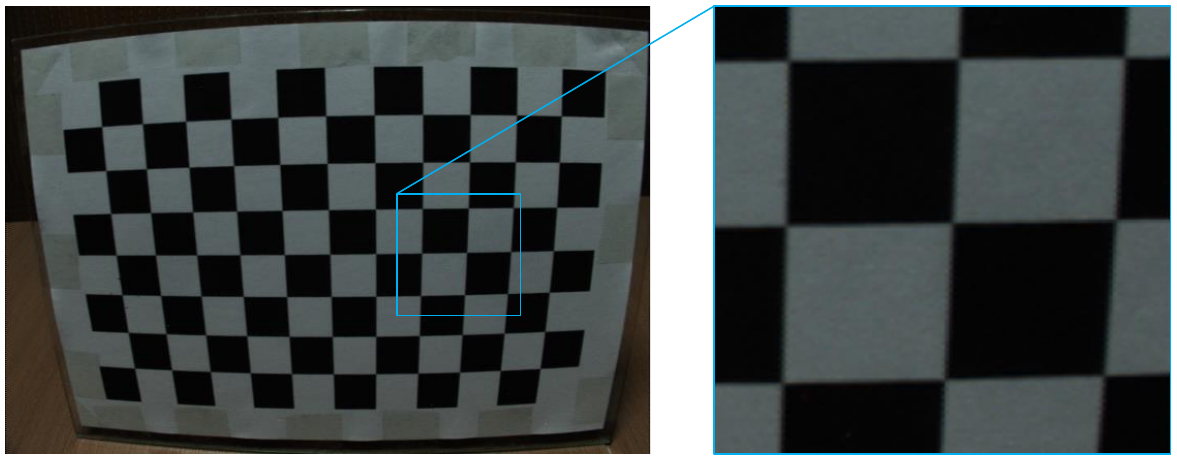}
\caption{Center view sub-aperture image extracted by method~\cite{dansereau2013decoding}.}
\label{fig:subCenter}
\end{figure}
\begin{figure}[bth]
\centering
\includegraphics[width=80mm]
{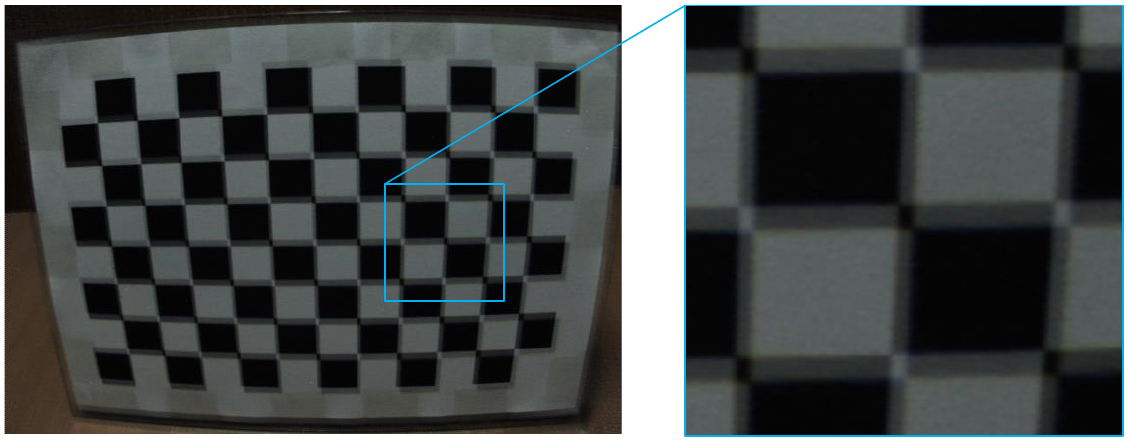}
\caption{Another sub-aperture image far from the optical axis, which is extracted by method~\cite{dansereau2013decoding}.}
\label{fig:subFar}
\end{figure}

As shown in the tables, the method in~\cite{dansereau2013decoding} performs well on dataset D-A, D-B and D-C with the criteria PR and PP, but with a relatively low performance on dataset P-A, P-B. The reason is that the corner locations are obtained from sub-aperture images. The sub-aperture image extraction method in~\cite{dansereau2013decoding} lose its efficiency for dataset P-A, P-B. Fig.~\ref{fig:subCenter} and Fig.~\ref{fig:subFar} are from a photo in dataset P-A, the center view sub-aperture image is still clear while for sub-aperture images far from the optical axis, some ghosts exit around the boundary of black areas and white areas.

For method in~\cite{bok2016geometric}, it performs well under PR metric while the performance drops significantly if measured by PP metric. The reason maybe that this method cannot ensure that the rays coming from those detected 2D corners corresponding to the same ideal 3D corner still converge to one single 3D point.

The proposed method performs well for both PR and PP metric on all datasets. It is due to that the proposed method extracts corner locations directly from raw images which avoid the sub-aperture images extraction procedure. Moreover, rather than several 2D projected corners, only one 'LF-points' is extracted for each 3D corner. It implies that all reconstructed rays of one 3D corner can still converge to one single 3D point.

Table~\ref{tab3} also shows that neither of the counterparts of the proposed ~\cite{dansereau2013decoding}~\cite{bok2016geometric} achieves satisfactory performance in terms of depth error. This is due to that the 'point to ray error' considered in~\cite{dansereau2013decoding} and the objective function 'line to ray error' considered in ~\cite{bok2016geometric} all belongs to lateral error. Errors in depth direction are not specially treated during the calibration thus making themselves less suitable for depth related applications. On the other hand, due to the parameters grouping and the separate calculation, the proposed calibration method achieves satisfactory performance for depth errors.

\section{Conclusion}

In this paper, detailed analysis of state-of-the-art projection model was performed to guide their usage in different applications. The analysis shows that a projection model can be interpreted in three expressions in terms of the correspondence between rays and pixels, 3D physical points and pixels and 3D physical points and 3D signal structure of the captured light field. Based on the analysis, parameters in the projection model can be further classified into direction parameter set and depth parameter set. Correspondingly, a two-step calibration method was proposed to deal with each group of the parameters. Systematic validations were conducted to evaluate the performance of the proposed calibration method. Experimental results show that the proposed method outperforms the state-of-the-art methods under various benchmark criteria.


\ifCLASSOPTIONcompsoc
  \section*{Acknowledgments}

\else
  \section*{Acknowledgment}
\fi

This work was supported by the National Science Foundation of China under Grant 61571337, 61601349 and 61801364.

\ifCLASSOPTIONcaptionsoff
  \newpage
\fi

\bibliographystyle{IEEEbib}
\bibliography{ref}

%

\end{document}